\documentclass[review, 3p, times]{elsarticle}

\biboptions{sort&compress}

\usepackage{amsmath}
\usepackage{amssymb}
\usepackage{setspace}
\usepackage{url}
\usepackage{placeins}
\usepackage{booktabs}
\usepackage{tabularx}
\usepackage[ruled,vlined,linesnumbered]{algorithm2e}
\usepackage{caption}
\usepackage{bm}
\usepackage{lineno}

\SetAlCapFnt{\footnotesize}
\SetAlCapNameFnt{\footnotesize\bfseries}

\DontPrintSemicolon
\SetAlgoNoEnd
\SetAlgoInsideSkip{smallskip}
\SetAlCapSkip{0.35em}
\SetNlSty{textbf}{}{:}

\SetKwInOut{Input}{Input}
\SetKwInOut{Output}{Output}
\SetKw{KwBreak}{break}
\SetKw{KwReturn}{return}
\SetKwFor{ForAll}{for all}{do}{}
\SetKwComment{Comment}{$\triangleright$\ }{}

\newcommand{\AlgProc}[1]{\textsc{#1}}

\def\tsc#1{\csdef{#1}{\textsc{\lowercase{#1}}\xspace}}
\tsc{WGM}
\tsc{QE}

\journal{Pattern Recognition}

\begin{document}

\begin{frontmatter}

\title{Data-Native Global Optimization for Big Data K-means Clustering}

\author[aff1]{Ravil Mussabayev\corref{cor1}\fnref{orcid-ravil}}
\ead{ravmus@gmail.com}
\ead{r.mussabayev@satbayev.university}

\author[aff1,aff2]{Rustam Mussabayev\fnref{orcid-rustam}}
\ead{ru.mussabayev@satbayev.university}
\ead{rustam@iict.kz}

\author[aff1]{Zukhra Yerdaliyeva}
\ead{z.yerdaliyeva@satbayev.university}

\author[aff1]{Kuldeyev Nursultan}
\ead{n.kuldeyev@satbayev.university}

\affiliation[aff1]{
  organization={AI Research Lab, Satbayev University},
  city={Almaty},
  postcode={050000},
  country={Kazakhstan}
}

\affiliation[aff2]{
  organization={Laboratory for Analysis and Modeling of Information Processes, Institute of Information and Computational Technologies},
  city={Almaty},
  postcode={050010},
  country={Kazakhstan}
}

\cortext[cor1]{Corresponding author}
\fntext[orcid-ravil]{ORCID: 0000-0003-1105-5990}
\fntext[orcid-rustam]{ORCID: 0000-0001-7283-5144}

\begin{abstract}
Big data clustering remains challenging: the Minimum Sum-of-Squares Clustering (MSSC) problem underlying K-means is NP-hard, and existing methods either reach poor local minima or require prohibitive metaheuristic hybrids. We target arbitrarily tall data: a fixed feature space may contain arbitrarily many, possibly infinitely many, observations, while the algorithm accesses only finite random samples. We propose Big-means++, a simple algorithm achieving scalability and global-search quality by systematically curating inputs to MSSC optimization on big data. It orchestrates local K-means refinements into a data-native global search for big data clustering. Rather than optimizing the full-data MSSC objective, Big-means++ traverses sample-induced surrogate landscapes. Each sample defines a cheap, distinct empirical MSSC approximation with a perturbed local-optimum structure, turning sample-to-sample variation into a global-search mechanism. Unlike vanilla Big-means, a flowing-incumbent strategy propagates centroid state across empirical landscapes through K-means refinements on fresh samples without rollback to a best-so-far solution. This increases mobility and favors stable, high-quality configurations across multiple approximations of the full-data structure. Moreover, a new shaking mechanism varies sample size in a geometric sequence, broadening the surrogate landscapes explored across multiple resolution scales, accounting for cluster imbalance, and improving solution quality. A competitive multi-agent system asynchronously explores independent sampled landscapes, transforming diverse stochastic trajectories into collective search intelligence. Automatic convergence detection stops each agent after attaining a high-quality solution but before further search risks degrading it, while providing a universal speed-quality control. Experiments on 22 datasets against 11 competing algorithms demonstrate the effectiveness, efficiency, and robustness of Big-means++.
\end{abstract}


\end{frontmatter}


\section{Introduction}
\label{sec:introduction}

Cluster analysis is a fundamental unsupervised learning tool used to discover structural patterns across diverse domains~\citep{Ezugwu2022Survey}. While broad surveys emphasize the role of clustering across modern machine-learning applications, a recent K-means-focused review shows that K-means variants remain central in large-scale and big-data clustering research~\citep{Ikotun2023KMeansReview}. At the heart of this field lies the Minimum Sum-of-Squares Clustering (MSSC) problem, commonly approached via the standard K-means algorithm~\citep{Forgy1965Cluster}. While K-means is fast and widely adopted, MSSC is known to be non-convex~\citep{Cuong2020MSSC} and NP-hard~\citep{Aloise2009MSSC}. In the era of Big Data, where datasets routinely span millions of high-dimensional records, these challenges are severely amplified~\citep{Mussabayev2023BigMeans}. Consequently, standard iterative algorithms are highly susceptible to becoming trapped in suboptimal local minima~\citep{Jain2010DataClustering}. Furthermore, the existing methods typically require full passes over the data, and even a handful of such iterations exhausts the available computational budget, while still failing to approach globally optimal solutions~\citep{Sculley2010WebScale}. The traditional remedy for escaping local optima involves integrating K-means with complex global optimization metaheuristics---such as differential evolution~\citep{Mansueto2021MDEClust}, genetic algorithms~\citep{Gribel2019HGMeans}, or non-smooth optimization~\citep{Karmitsa2025BigClust}. However, these hybrid approaches are notoriously difficult to understand and implement, discouraging broad adoption~\citep{Mladenovic2022LessIsMore}; moreover, their excessive computational overhead renders them impractical for large-scale clustering. Paradoxically, these complex methods frequently fail to match the clustering accuracy of far simpler data-driven alternatives~\citep{Mussabayev2023BigMeans,Mussabayev2024CompAnal}.

Formally, given a dataset $X = \{x_1, \ldots, x_m\} \subset \mathbb{R}^n$ of $m$ points, MSSC~\citep{Lloyd1982Least} seeks $k$ centroids $C = (c_1, \ldots, c_k) \in \mathbb{R}^{n \times k}$ that minimize the total intra-cluster variance:
\begin{equation}
    \min_{C} \; f(C, X) = \sum_{i=1}^{m} \min_{j=1,\ldots,k} \|x_i - c_j\|^2,
    \label{eq:mssc}
\end{equation}
where $\|\cdot\|$ denotes the Euclidean norm. The centroids $C$ uniquely induce a partition $X = X_1 \cup \cdots \cup X_k$, where cluster $X_j$ contains all points for which $c_j$ is the nearest centroid; thus, minimizing~\eqref{eq:mssc} simultaneously maximizes intra-cluster cohesion and inter-cluster separation. MSSC always possesses finite global solutions and stable, predictable behavior under data perturbations~\citep{Cuong2020MSSC}, and global minimizers are known to yield the most faithful representations of the underlying cluster structure~\citep{Gribel2019HGMeans}, yet the highly non-convex landscape makes them notoriously difficult to attain. When the data stream is effectively unbounded, i.e., $m \to \infty$, formulation~\eqref{eq:mssc} extends to the Minimum Sum-of-Squares Clustering of Infinitely Tall Data (MSSC-ITD) problem~\citep{Mussabayev2024HPClust}, which additionally requires algorithms capable of global search without ever materializing the full dataset in memory.

Big-means~\citep{Mussabayev2023BigMeans} reframes large-scale clustering by replacing the classical ``optimize on the full dataset'' paradigm with a sequence of fast, full local K-means optimizations on randomly sampled subproblems. Rather than iterating over the entire dataset, it treats it as an infinite source of cheap data samples. At each iteration, a small random subset $S$ of size $s \ll m$ is drawn without replacement, and K-means is run to convergence starting from the best centroid state attained so far (with K-means++ used for initial seeding and dynamic repair of degenerate centers). This process repeats until a time or iteration budget is exhausted. Because a random sample $S$ is a statistically representative snapshot of the full dataset, its objective landscape $f(\cdot, S)$ provides a slightly perturbed distribution of the true local optima. Iteratively optimizing these subproblems drives the centers toward solutions that remain stable across many random views, while enabling the algorithm to bypass poor local minima---whether of the full-data objective or of individual samples. The resulting trajectory can be viewed as a descending walk through local optima of sample-induced MSSC landscapes, where landscape variability naturally prevents stagnation. The cost per iteration reduces from $\mathcal{O}(m \cdot k \cdot n \cdot \tau)$ (standard K-means' complexity) to $\mathcal{O}(s \cdot k \cdot n \cdot \tau)$ (where $\tau$ is the number of K-means sweeps), decoupling computational complexity from $m$. This allows many more improvement attempts per unit time, dramatically enhancing the exploration-exploitation trade-off.

The deepest idea in Big-means is that the sample performs two jobs simultaneously. First, it is a \emph{decomposition device} that makes K-means cheap enough for big data. Second, it is a \emph{landscape-variation device}: every new sample slightly shifts the distribution of local minima and how attractive they are. Here, sample-to-sample variability is not noise to be eliminated; it is the natural global search operator. Following a ``less is more'' philosophy~\citep{Mladenovic2022LessIsMore}, instead of adding a heavy global-search mechanism, the algorithm gains robust neighborhood variation almost for free. This differs from restart-based improvements of K-means, where careful initialization and repeated runs can substantially improve solution quality~\citep{Franti2019KMeansInitRepeats} but do not change the underlying full-data local-search paradigm. Furthermore, unlike mini-batch methods~\citep{Sculley2010WebScale} that make small stochastic updates, each move in Big-means++ is \emph{strong}: running a full, real local K-means procedure on each sample ensures every update is a substantial optimization step on the sample's objective landscape rather than just a noisy incremental correction.

Despite its promise, Big-means has two fundamental limitations. First, it requires manual per-dataset tuning of key parameters---most notably the sample size and the time or iteration budget---which severely hinders its adaptability and practical adoption~\citep{Karmitsa2025BigClust}. Second, it employs an elitist acceptance rule: any sample-based refinement that fails to improve upon the best-so-far solution is discarded. As we demonstrate experimentally, this severely restricts the mobility of the incumbent centroids, squandering the directional signal embedded in non-improving---yet still data-informed---refinements.

Several derivative methods have since been proposed within the Big-means framework~\citep{Mussabayev2024Superior,Mussabayev2024HPClust,Mussabayev2025BiModalClust}, each improving upon specific aspects but inheriting both limitations. Notably, results in~\citep{Mussabayev2024Superior} showed that varying the sample size during search improves solution quality and computational efficiency even relative to repeated runs with manually optimized fixed sizes, while also reducing tuning effort. This suggests that sample size is not merely a hyperparameter to be selected, but an additional degree of freedom for shaking sample-induced landscapes, motivating a principled redesign of the framework.

Below, we summarize our main contributions---both broadly and with respect to our previous work:
\begin{enumerate}[1.]
	\item \textbf{Data-Native Optimization Concept:} We demonstrate that the Variable Landscape Search (VLS)~\citep{Mussabayev2025VLS} and ``less is more''~\citep{Mladenovic2022LessIsMore} principles hold: restricting optimization to the input-curation modality alone, without any metaheuristic hybridization, suffices to outperform significantly more complex competing methods for an NP-hard problem.
    \item \textbf{Flowing Incumbent for Deeper Exploration:} We simplified the global search process by removing the greedy acceptance rule. Unconditional incumbent updates allow the algorithm to dynamically ride the shifting sample-induced landscapes, leading to significantly deeper exploration and lower final objective values.
	\item \textbf{Shaking via Sample Size Variation:} We introduce a geometric sequence of sample sizes---a ``ladder''---that navigates multiple resolution scales during the search, accounting for cluster imbalance and steering the balance between diversification and intensification. Essentially, this serves as a new form of shaking for MSSC global search, in which geometric variation of the sample size exposes the search to qualitatively different surrogate landscapes and ultimately yields better solutions.
	\item \textbf{Collective Multi-Agent Search}: We organize the search as a competitive population of autonomous agents that independently traverse sampled landscapes and preserve complementary stochastic trajectories. A final common-landscape evaluation transforms this decentralized diversity into a collective decision, enabling broader global exploration than any single trajectory alone.
	\item \textbf{Plug-and-Play Adaptiveness:} Adaptiveness is not merely a utility for external hyperparameter tuning, but an integral optimization principle. Under the flowing-incumbent strategy, a data-adaptive per-agent convergence detector stops each trajectory when continued landscape perturbations cease to produce meaningful progress, limiting the risk that unnecessary search degrades an already strong solution. It thereby regulates useful search effort, allocates computation according to trajectory difficulty, and eliminates dataset-specific budget tuning while retaining universal speed-quality controls.
	\item \textbf{Standardized Benchmarking Framework:} We developed a reproducible benchmarking suite with rigorous statistical analysis tools (available upon request). To ensure fair, low-level performance comparisons under identical yet highly varied conditions, we reimplemented 11 competing algorithms in C++---retaining Fortran only for nonsmooth methods that rely on extensive native codebases---and evaluated Big-means++ comprehensively across 22 real-world datasets spanning diverse scales.
\end{enumerate}

\section{Related Work}
\label{sec:related_work}

Research on big-data MSSC has followed three main directions. The first seeks scalability mainly through cheaper approximations of the classical K-means pipeline. This includes stochastic or online updates such as MiniBatchKMeans~\citep{Sculley2010WebScale}, scalable seeding schemes such as K-means||~\citep{Bahmani2012ScalableKMeansPP}, divide-and-merge K-means variants such as BDCSM~\citep{Alguliyev2020BDCSM}, and data-reduction techniques based on weighted summaries or coresets~\citep{Bachem2017LightweightCoresets,Bachem2017OneShotCoresets}. Capo et al.~\citep{Capo2020TallKMeans} specifically address tall data through a recursive parallel approximation to K-means, but the method remains primarily an acceleration strategy rather than a mechanism for traversing sample-induced objective landscapes. The scalable-seeding line has also been strengthened theoretically: Makarychev et al.~\citep{Makarychev2020ImprovedKMeansPP} improve approximation guarantees for both K-means++ and its parallel variant. These methods reduce passes over the data, communication, memory footprint, or initialization cost, but they typically preserve the local-search character of K-means on a largely fixed objective landscape and therefore prioritize throughput over deeper global exploration.

The second direction improves solution quality by adding explicit global-search machinery on top of K-means. Earlier Pattern Recognition work illustrates the breadth of K-means-centered quality improvements: J-means introduces local-search moves for improving MSSC solutions~\citep{Hansen2001JMeans}, whereas I-k-means-+ iteratively enhances the K-means procedure itself~\citep{Ismkhan2018IKMeans}. Representative examples of heavier global-search methods include population-based or memetic methods such as HG-means~\citep{Gribel2019HGMeans} and MDEClust~\citep{Mansueto2021MDEClust}, random-swap style search such as DRSMeans~\citep{Kozbagarov2024DRSMeans}, and the nonsmooth optimization line represented by LMBM-Clust~\citep{Karmitsa2018LMBMClust}, BigClust~\citep{Karmitsa2025BigClust}, and Clust-Splitter~\citep{Lampainen2026ClustSplitter}. These methods can locate substantially deeper local minima, but they do so by introducing additional algorithmic ingredients---e.g., recombination, stochastic swaps, auxiliary formulations, or sequential bundle optimization---which usually increases implementation complexity and computational overhead. At the same time, Kalczynski et al.~\citep{Kalczynski2022LessIsMoreMSSC} show that relatively simple MSSC algorithms can remain competitive, reinforcing the motivation for designs that gain search power without excessive algorithmic machinery.

A third direction, closer to our work, uses data decomposition itself as the main optimization mechanism. The original Big-means~\citep{Mussabayev2023BigMeans} showed that repeatedly solving K-means to convergence on fresh random samples can use sampling not only as a scalability device but also as a global-search operator: each sample induces a slightly different MSSC landscape, and moving across these landscapes helps escape poor local minima without heavy hybridization. Recent extensions enriched this family with competitive sample-size adaptation~\citep{Mussabayev2024Superior}, competitive multi-agent search strategies for MSSC-ITD~\citep{Mussabayev2024HPClust}, and explicit bimodal hybridization with neighborhood search~\citep{Mussabayev2025BiModalClust}. Viewed through the Variable Landscape Search perspective~\citep{Mussabayev2025VLS}, these methods optimize by traversing a family of related sample-induced landscapes rather than by searching only one static full-data landscape.

Big-means++ is not merely another variant of this family; it reformulates sample-based MSSC search as a multi-resolution landscape-continuation process. Starting from the Big-means insight that random samples provide computationally cheap perturbations of the MSSC landscape~\citep{Mussabayev2023BigMeans}, the proposed method introduces three methodological shifts. First, the incumbent becomes a mobile, ever-flowing state propagated through successive sample-induced landscapes. Second, sample size is promoted from a fixed hyperparameter to an active search variable: geometric variation of its scale changes the resolution and local-optimum structure of the surrogate landscapes, broadening the landscape family traversed and regulating diversification versus intensification. Third, independent search trajectories are organized as a competitive multi-agent population. By preventing premature consensus during exploration and comparing all final agent states on a common sampled landscape, Big-means++ converts distributed search diversity into collective optimization intelligence. Convergence detection complements these mechanisms by adapting the duration of each trajectory, stopping it near the point where useful landscape-driven improvement gives way to unproductive movement. Besides protecting solution quality, this autonomously allocates search effort and removes the need for instance-specific budget tuning.

Thus, Big-means++ operationalizes the Variable Landscape Search view for big-data MSSC: global exploration is achieved not by adding an external metaheuristic layer, but by organizing both the sequence and resolution of the data-induced landscapes through which K-means moves~\citep{Mussabayev2025VLS}. The method concentrates search power in native degrees of freedom of big data itself---the sampled observations, the sample scale, and the trajectory connecting successive sample-based optima.

\section{The Big-means++ Algorithm}
\label{sec:algorithm}

\subsection{Algorithm Overview}

Figure~\ref{fig:bigmeans-pp-diagram} illustrates the overall structure of the proposed method, and Algorithm~\ref{alg:big-means-plus-plus} gives its formal description.

\par\bigskip
\noindent\begin{minipage}{\linewidth}
	\centering
	\includegraphics[width=\linewidth]{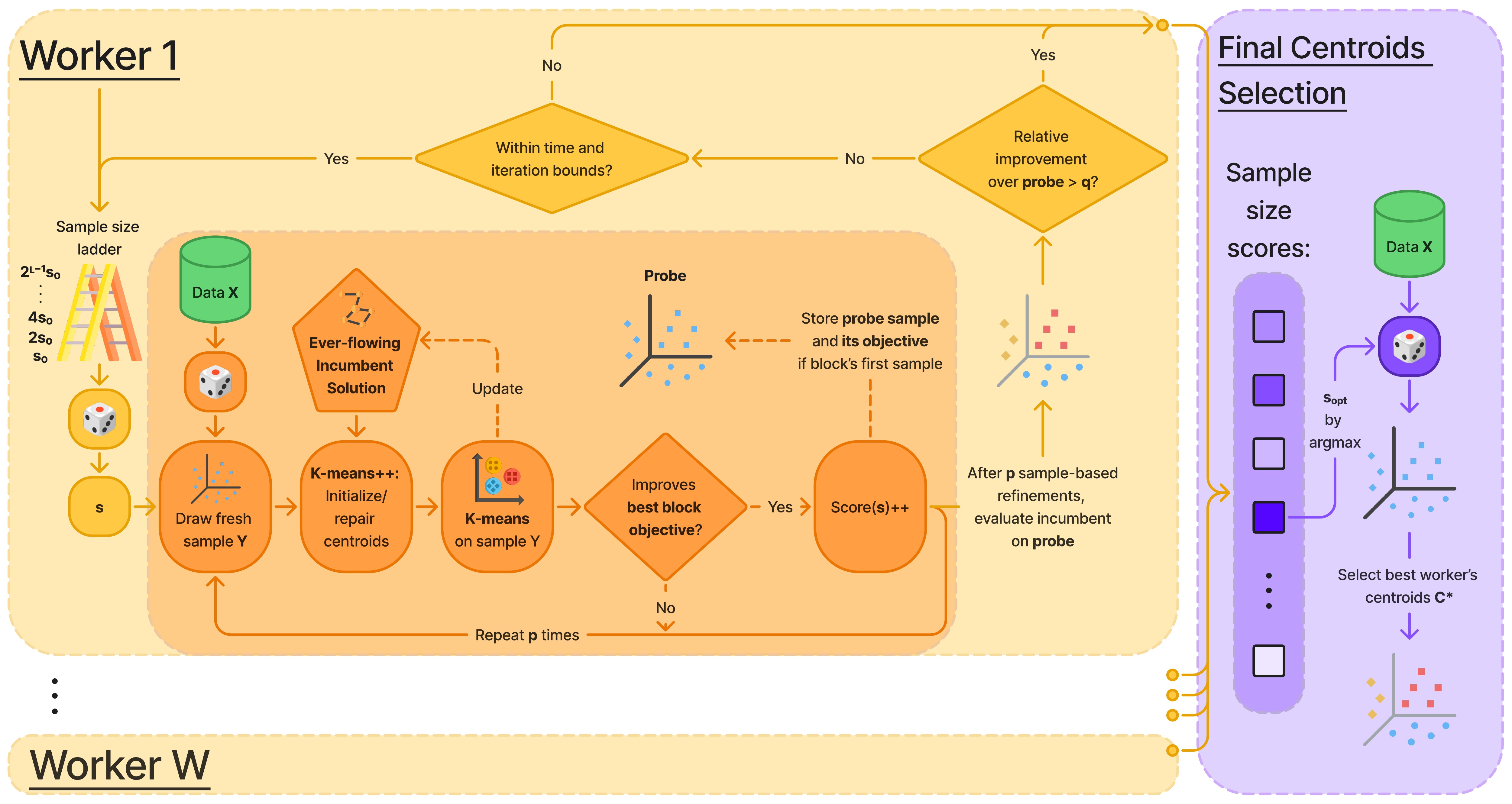}
	\captionof{figure}{Schematic diagram of the Big-means++ algorithm.}
	\label{fig:bigmeans-pp-diagram}
\end{minipage}
\par\bigskip

\begin{algorithm}[H]
	\caption{Big-means++}
	\label{alg:big-means-plus-plus}
	\footnotesize
	\setstretch{1.05}
	
	\Input{Dataset $X\subset\mathbb{R}^n$; number of clusters $k$;
		block size $p$; probe tolerance $q$; budgets $I_{\max}$ and $T_{\max}$;
		local K-means parameters; sample-size ladder parameters $s_0$ and $L$.}
	
	\Output{Centroids $C^\star$, objective value $f^\star$, assignments $a^\star$,
		and selected sample size $s_{\mathrm{opt}}$.}
	
	Build a geometric sample-size ladder
	$\mathcal{S}=\{s_0,2s_0,4s_0,\ldots\}$, capped by $|X|$
	\Comment*[r]{multi-resolution landscape shaking}
	
	For each $s\in\mathcal{S}$, set $\mathrm{score}(s)\gets0$\;
	Let $W$ be the number of autonomous search agents
	\Comment*[r]{implemented as parallel workers}
	
	\BlankLine
	
	\ForAll{agents $w=1,\ldots,W$ \textnormal{\textbf{in parallel}}}{
		$C_w\gets(\varnothing_1,\ldots,\varnothing_k)$
		\Comment*[r]{current flowing centroids}
		
		\While{time and iteration budgets are not exhausted}{
			Draw a sample size $s$ uniformly from $\mathcal{S}$\;
			$\mathrm{improved}\gets0$; $\mathrm{probe}\gets\varnothing$\;
			
			\For{$r=1,\ldots,p$}{
				Draw a fresh sample $Y\subset X$ of size $s$\;
				Initialize or repair missing centroids in $C_w$ by K-means++ on $Y$\;
				Refine $C_w$ by local K-means on $Y$
				\Comment*[r]{centroids keep flowing}
				
				\If{this refinement improves the current block objective}{
					$\mathrm{improved}\gets\mathrm{improved}+1$\;
					\lIf{$\mathrm{probe}=\varnothing$}{store $Y$ and its objective as the block probe}
				}
			}
			
			Increase $\mathrm{score}(s)$ by $\mathrm{improved}$\;
			
			\If{a block probe was stored}{
				Evaluate the final block centroids on the stored probe sample\;
				\If{their relative improvement over the probe objective is at most $q$}{
					\KwBreak
					\Comment*[r]{stop before unproductive trajectory drift}
				}
			}
		}
	}
	
	\BlankLine
	
	$s_{\mathrm{opt}}\gets\arg\max_{s\in\mathcal{S}}\mathrm{score}(s)$
	\Comment*[r]{most productive observed scale}
	Draw a common evaluation sample $Y_{\mathrm{eval}}\subset X$ of size $s_{\mathrm{opt}}$\;
	
	\For{$w=1,\ldots,W$}{
		Repair missing centroids in $C_w$ on $Y_{\mathrm{eval}}$, if necessary\;
		Evaluate $C_w$ on $Y_{\mathrm{eval}}$\;
	}
	
	Select the agent $w^\star$ with the best objective on $Y_{\mathrm{eval}}$\;
	$C^\star\gets C_{w^\star}$\;
	$(f^\star,a^\star)\gets\AlgProc{Assign}(X,C^\star)$
	\Comment*[r]{single full-data pass}
	
	\KwReturn{$(C^\star,f^\star,a^\star,s_{\mathrm{opt}})$}
\end{algorithm}

The C++ source code of Big-means++ is publicly available at \url{https://github.com/rmusab/bigmeans-pp}.

The key operation of Big-means++ is a block of local search refinements on sample-induced objective landscapes. Within a block, the centroids are \emph{flowing}: after each sampled local K-means refinement, the resulting centroids are immediately used as the starting point for the next sample. The first improving sample in a block is stored as a convergence probe, providing a fixed reference against which subsequent movement is assessed. At the end of the block, the final centroids are evaluated on this probe. If their relative improvement is at most $q$, further landscape perturbations are deemed unlikely to yield sufficient benefit, and the agent stops before unnecessary continuation can erode the quality of its terminal state. In our implementation, the local K-means refinement is accelerated by Hamerly's distance bounds~\citep{Hamerly2010Making}. Because empty clusters can materially affect K-means behavior, degenerate-center repair has been studied as a methodological issue in its own right~\citep{Aloise2017DegenerateKMeans}; in Big-means++, missing centroids are repaired by K-means++ initialization~\citep{Arthur2007KmeansPP}.

To navigate varied landscapes, Big-means++ adopts a \emph{``flowing incumbent''} mechanism: the centroid state is passed unconditionally from one random sample to the next, without rolling back to a previous best solution. This non-elitist strategy acts as a Markov chain through the solution space. Each iteration slightly perturbs the centers (because the sample is different), and that perturbed state is always carried forward. Even a momentarily worse state is still informed by the data and can nudge the centers into a region from which the next sample leads to a strict improvement. Over a fixed time budget this explores far more of the solution space and consistently reaches lower objectives than the standard ``sticky'' strategy (greedy hill-climbing anchored to a fixed point), which is employed by many other clustering algorithms. The sticky update policy tends to produce premature convergence: the search quickly finds a local basin of attraction and spends the remainder of the time budget in futile attempts to escape it, causing many rejected iterations. In contrast, the flowing update turns the same wall-clock budget into genuine exploration, gaining mobility across distinct basins of attraction and implicitly pushing the trajectory toward configurations that are simultaneously good across multiple perturbations.

Varying sample identity at a fixed size perturbs the MSSC landscape within a single resolution, but confines the search to a comparatively narrow family of surrogate landscapes. Big-means++ therefore introduces sample size as a second data-native shaking coordinate through a \emph{geometric ladder of sample sizes}: $\{s_0,2s_0,4s_0,\dots,2^{L-1}s_0\}$, capped at $m$. Changing the sample size alters the statistical resolution and local-optimum structure of the sampled objective: small samples produce stronger, cheaper perturbations that promote broad exploration, whereas large samples provide more stable descent signals and better coverage of small clusters. Traversing these scales broadens the family of landscapes explored, accounts for cluster imbalance, and regulates diversification versus intensification. For each block of $p$ iterations, the algorithm uniformly selects one ladder level and records how often the sample objective improves.

Big-means++ organizes global exploration as a competitive multi-agent system. Each autonomous agent maintains its own flowing centroid state and independently traverses a private sequence of sampled landscapes and resolution scales. Because agents do not exchange incumbents during search, they avoid premature consensus and preserve complementary hypotheses about the cluster structure. Their benefit is therefore not limited to parallel speedup: the population collectively covers a broader range of stochastic search trajectories than any single agent can explore. After convergence, all agent states compete on a common sampled landscape, converting decentralized trajectory diversity into a single collective decision. In our implementation, agents are realized as independent OpenMP workers, requiring no communication or synchronization during search.

Under a flowing incumbent, additional search is not automatically beneficial: because every sampled refinement is accepted, an agent that has reached a strong state may subsequently drift to a weaker one. The probe-based convergence detector therefore acts as an optimization driver, not merely a computational shortcut. Each agent stores the first improving sample within a block and evaluates its final block centroids on that same probe. If their relative improvement does not exceed $q$ (typically $0.001$), the trajectory stops near the transition from productive movement to diminishing or potentially harmful continuation. This mechanism gives difficult trajectories more opportunity to improve, terminates exhausted trajectories early, and eliminates manual per-dataset budget tuning at the cost of only one assignment pass on a small sample.

In Big-means++, each local K-means call uses Hamerly's algorithm instead of Lloyd's. By maintaining distance bounds, it heavily reduces distance evaluations without affecting correctness.

Finally, because agents explore different sample sequences and resolution scales, their terminal objectives are not directly comparable. Big-means++ therefore completes the collective search with a common-landscape evaluation: all final agent states are assessed on the same fresh sample, providing a shared competitive arena from which the strongest collective candidate is selected. It identifies $s_{opt}$, the sample scale associated with the most improvements during search, and uses a fresh common sample of this size for the comparison. This unbiased, apples-to-apples evaluation guarantees the best candidate is chosen based on a representative validation criterion. Crucially, the full dataset is accessed only once at the very end to evaluate and report the global quality of the winning solution. The optimization phase itself operates entirely free from full-dataset scans, cementing the method's scalability.

\subsection{Choice of Universal Parameters}

Big-means++ has four main universal parameters: the block size $p$, the convergence tolerance $q$, the initial sample size $s_0$, and the ladder depth $L$. These parameters are not tuned delicately for each dataset; rather, they define a robust default search regime.

In all experiments, we use $p=10$. This value provides a good balance between allowing the incumbent solution to explore the current scale and checking convergence frequently enough. Its robustness is further examined in the sensitivity analysis reported in the experimental section.

The convergence tolerance $q$ is the primary universal control over how long an agent continues after useful movement begins to diminish. Smaller values demand stronger evidence of convergence, permitting deeper search but increasing the opportunity for further trajectory drift and computation; larger values stop more aggressively, protecting strong intermediate states and reducing runtime. Thus, $q$ provides a universal quality-time control without requiring a separately optimized budget for each dataset. It is typically chosen from the range $[0.001, 0.1]$. In our experiments, we set $q \in \{0.001, 0.05\}$ for the quality-oriented mode. For Big-means++$_\mathrm{Fast}$, the efficiency-oriented variant, we adopt a more aggressive tolerance $q_{\mathrm{fast}} \in \{0.05, 0.1\}$. The resulting per-dataset selections are summarized in Table P1 and Tables S1--S22 of the Supplementary Material.

By default, the ladder is initialized at $s_0 = \min\{m, 2048\}$. For strongly imbalanced datasets, however, small initial samples may fail to capture rare clusters. In such cases, $s_0$ should in principle be chosen to guarantee rare-cluster coverage: specifically, it must be large enough so that a random sample captures the smallest practically relevant cluster with high probability. In practice, however, it suffices to set $s_0 \in \left\{ \operatorname{pow}_2\!\left(\frac{m}{4}\right),\, \operatorname{pow}_2\!\left(\frac{m}{2}\right) \right\}$, where $\operatorname{pow}_2(\cdot)$ denotes rounding to the nearest power of two. This yields a ladder whose lower bound supports rare-cluster coverage while preserving a sufficiently broad range of landscape resolutions for effective shaking.

For the quality-oriented variant of Big-means++, we fix the ladder depth at $L = 9$ across all datasets, as this value spans a sufficiently broad range of landscape resolutions while remaining computationally tractable for the full range of dataset scales considered. Big-means++$_\mathrm{Fast}$ uses the same values of $p$ and $s_0$, but replaces $q$ and $L$ with $q_{\mathrm{fast}}$ and $L_{\mathrm{fast}} \le L$. In our experiments, we set $L_{\mathrm{fast}} = L = 9$ on all datasets except the largest ones: $L_{\mathrm{fast}} = L = 5$ for CORD-19 Embeddings, $L_{\mathrm{fast}} = 6$ for HEPMASS, and $L_{\mathrm{fast}} = 3$ for Gisette. Hence, Big-means++$_\mathrm{Fast}$ combines a more aggressive convergence test with a shorter ladder, exploring the same lower sample scales while avoiding the most expensive upper levels.

\subsection{Computational Complexity}

The computational advantage of Big-means++ is that its search phase is performed on
samples rather than on the full dataset. If a local K-means refinement on a sample of
size $s$ requires $\tau$ iterations, its standard worst-case cost is
$\mathcal{O}(\tau s k n)$. Hamerly's acceleration reduces the practical number of
distance computations, but does not change this worst-case order.

Although the per-iteration cost is simple, Lloyd-style K-means can require
exponentially many iterations in worst-case constructions even in two
dimensions~\citep{Vattani2011ExponentialKMeans}. This further motivates the
blockwise convergence detector used by Big-means++ to regulate sampled local-search
trajectories in practice.

Let $W$ be the number of autonomous agents, implemented as parallel workers, $p$ the
block size, $B$ the average number of executed blocks per agent before convergence,
and $\bar{s}$ the average sample
size drawn from the sample-size ladder. Ignoring lower-order repair and bookkeeping
costs, the sampled search phase requires
\[
    \mathcal{O}\!\left(W B p\,\tau\,\bar{s}\,k n\right)
\]
total work. The final agent selection on the common validation sample of size
$s_{\mathrm{opt}}$ costs $\mathcal{O}(W s_{\mathrm{opt}} k n)$, while the final
assignment of all data points to the selected centroids requires one full-data pass,
that is, $\mathcal{O}(mkn)$. Thus, the overall work is
\[
    \mathcal{O}\!\left(
    W B p\,\tau\,\bar{s}\,k n
    + W s_{\mathrm{opt}} k n
    + mkn
    \right).
\]

Since agents search independently, their computational realization is embarrassingly parallel. Hence,
up to load imbalance and hardware overheads, the wall-clock cost is governed by one
agent's sampled trajectory plus the final validation and full-data assignment passes.
For fixed universal parameters $W$, $p$, and $L$, the optimization phase is therefore
decoupled from $m$; the only unavoidable linear dependence on the full dataset size is
the final reporting pass. This property makes Big-means++ particularly suitable for
big-data MSSC and MSSC-ITD settings. The memory usage is
$\mathcal{O}(mn + Wkn + s_{\max}n)$ when the dataset is stored in memory, or
$\mathcal{O}(Wkn+s_{\max}n)$ apart from the external data source in streaming-style
settings.

\section{Experimental Methodology}
\label{sec:methodology}

This section describes the experimental protocol used to evaluate Big-means++.
The empirical study is designed around the MSSC objective~\eqref{eq:mssc} itself: every method
is asked to produce $k$ centroids, the returned centroids are evaluated on the
full dataset by one final assignment pass, and solution quality is measured by
the resulting sum of squared distances. The experimental outcomes are reported and discussed in Section~\ref{sec:results}.
The Supplementary Material provides the full numerical companion to this protocol:
Table P1 gives the parameter-selection policy for each benchmark algorithm,
Tables S1--S22 report detailed per-dataset results, and Figures S1--S26 provide
additional performance-profile, category, and variability analyses.

\subsection{Benchmark Datasets}

We use 22 publicly available real-world datasets that have also been used in
recent large-scale MSSC studies~\citep{Karmitsa2018LMBMClust,Mussabayev2023BigMeans,Karmitsa2025BigClust,Lampainen2026ClustSplitter}.
The collection deliberately spans diverse regimes: high-dimensional datasets
with only thousands of points, low-dimensional datasets with hundreds of
thousands or millions of points, and datasets that are large in both $m$ and
$n$. All datasets contain numerical features and are used in their non-normalized
form, matching the setting of the main rival studies whenever possible. The
number of data points ranges from $15{,}112$ to $10{,}500{,}000$, and the number
of features ranges from $2$ to $5000$.

Table~\ref{tab:datasets} summarizes the benchmark data. The rows are sorted by
decreasing problem size $m\times n$, where $m$ denotes the number of data points
and $n$ denotes the number of features. This ordering is also used in
Tables S1--S22 of the Supplementary Material, where each dataset is reported
across all eight values of $k$ together with per-$k$ summary statistics over
independent runs and the Big-means++ parameter settings used for that dataset.

\begin{table}[t]
	\caption{Benchmark datasets, sorted by decreasing $m\times n$.}
	\label{tab:datasets}
	\centering
	\footnotesize
	\setlength{\tabcolsep}{3pt}
	\begin{tabularx}{\textwidth}{@{}>{\raggedright\arraybackslash}p{0.24\textwidth}>{\raggedright\arraybackslash}Xrrr@{}}
		\toprule
		Dataset & URL & $m$ & $n$ & $m\times n$ \\
		\midrule
		CORD-19 Embeddings & \url{https://www.kaggle.com/datasets/allen-institute-for-ai/CORD-19-research-challenge} & 1,056,660 & 768 & 811,514,880 \\
		HEPMASS & \url{https://archive.ics.uci.edu/dataset/347/hepmass} & 10,500,000 & 28 & 294,000,000 \\
		US Census Data 1990 & \url{https://archive.ics.uci.edu/dataset/116/us+census+data+1990} & 2,458,285 & 68 & 167,163,380 \\
		Gisette & \url{https://archive.ics.uci.edu/dataset/170/gisette} & 13,500 & 5000 & 67,500,000 \\
		Music Analysis & \url{https://archive.ics.uci.edu/dataset/386/fma+a+dataset+for+music+analysis} & 106,574 & 518 & 55,205,332 \\
		BitcoinHeist & \url{https://archive.ics.uci.edu/dataset/526/bitcoinheistransomwareaddressdataset} & 2,916,697 & 8 & 23,333,576 \\
		Protein Homology & \url{https://www.kdd.org/kdd-cup/view/kdd-cup-2004/Data} & 145,751 & 74 & 10,785,574 \\
		MiniBooNE Particle Identification & \url{https://archive.ics.uci.edu/dataset/199/miniboone+particle+identification} & 130,064 & 50 & 6,503,200 \\
		Covertype & \url{https://archive.ics.uci.edu/dataset/31/covertype} & 581,012 & 10 & 5,810,120 \\
		MFCCs for Speech Emotion Recognition & \url{https://www.kaggle.com/datasets/cracc97/features} & 85,134 & 58 & 4,937,772 \\
		ISOLET & \url{https://archive.ics.uci.edu/dataset/54/isolet} & 7,797 & 617 & 4,810,749 \\
		Sensorless Drive Diagnosis & \url{https://archive.ics.uci.edu/dataset/325/dataset+for+sensorless+drive+diagnosis} & 58,509 & 48 & 2,808,432 \\
		Online News Popularity & \url{https://archive.ics.uci.edu/dataset/332/online+news+popularity} & 39,644 & 58 & 2,299,352 \\
		Gas Sensor Array Drift & \url{https://archive.ics.uci.edu/dataset/224/gas+sensor+array+drift+dataset} & 13,910 & 128 & 1,780,480 \\
		Range Queries Aggregates & \url{https://archive.ics.uci.edu/dataset/493/query+analytics+workloads+dataset} & 200,000 & 7 & 1,400,000 \\
		3D Road Network & \url{https://archive.ics.uci.edu/dataset/246/3d+road+network+north+jutland+denmark} & 434,874 & 3 & 1,304,622 \\
		KEGG Metabolic Relation Network (Directed) & \url{https://archive.ics.uci.edu/dataset/220/kegg+metabolic+relation+network+directed} & 53,413 & 20 & 1,068,260 \\
		Skin Segmentation & \url{https://archive.ics.uci.edu/dataset/229/skin+segmentation} & 245,057 & 3 & 735,171 \\
		Shuttle Control & \url{https://archive.ics.uci.edu/dataset/148/statlog+shuttle} & 58,000 & 9 & 522,000 \\
		EEG Eye State & \url{https://archive.ics.uci.edu/dataset/264/eeg+eye+state} & 14,980 & 14 & 209,720 \\
		Pla85900 & \url{https://softlib.rice.edu/pub/tsplib/tsp/pla859.tsp.gz} & 85,900 & 2 & 171,800 \\
		D15112 & \url{https://github.com/mastqe/tsplib/blob/master/d15112.tsp} & 15,112 & 2 & 30,224 \\
		\bottomrule
	\end{tabularx}
\end{table}

\subsection{Compared Algorithms}

The comparison contains 11 competitor algorithms and five Big-means++-family
variants. The competitors are divided into three groups.

The first group consists of classical or lightweight baselines: Forgy-initialized
K-means~\citep{Forgy1965Cluster}, MiniBatchKMeans from the Scikit-learn library~\citep{Sculley2010WebScale}, and BDCSM (Big Data Clustering on a Single Machine), a parallel batch K-means method~\citep{Alguliyev2020BDCSM}. These
methods represent simple full-data, mini-batch, and batch-merge strategies.

The second group contains recent optimization-oriented MSSC algorithms:
Clust-Splitter~\citep{Lampainen2026ClustSplitter},
Big-Clust~\citep{Karmitsa2025BigClust}, and
LMBM-Clust~\citep{Karmitsa2018LMBMClust}, all based on nonsmooth optimization;
MDEClust~\citep{Mansueto2021MDEClust}, a memetic differential-evolution method;
and DRS-means~\citep{Kozbagarov2024DRSMeans}, a density-based random-swap
method. These algorithms are included because they explicitly target high
quality MSSC solutions rather than only fast approximate clustering.

The third group contains direct predecessors from the Big-means line:
Big-means-Inn, the original Big-means algorithm with inner parallelism
from~\citep{Mussabayev2023BigMeans}; Big-means-Com, the competitive multi-agent
variant from~\citep{Mussabayev2024HPClust}; and
BigOptimaS3~\citep{Mussabayev2024Superior}, which augments competitive
Big-means with stochastic sample-size optimization. This group is essential for
isolating how much of the improvement comes from the new Big-means++ design
rather than from the broader Big-means decomposition principle alone.

Also, there are the quality-oriented Big-means++ algorithm, the faster Big-means++$_\mathrm{Fast}$ variant, and three ablations: one that removes the probe-based convergence detector, one that removes the geometric sample-size ladder by fixing a single sample size, and one that removes the flowing-incumbent mechanism by accepting only improving candidate states. These ablated variants are used to evaluate the contribution of the main algorithmic components.

\subsection{Benchmark Protocol and Parameter Settings}

Each dataset is clustered for $k \in \{2,3,4,5,10,15,20,25\}$, giving $22 \times 8 = 176$ benchmark instances in total. For stochastic algorithms, independent runs use the run index as the random seed, ensuring full reproducibility across all algorithms when the same seeds and number of agents/workers are used. Following the dataset configuration used by the testbed, the
four largest or most expensive benchmark datasets in terms of computation
policy---CORD-19 Embeddings, HEPMASS, US Census Data 1990, and Gisette---use
five independent runs, while the remaining datasets use ten independent runs.
Deterministic algorithms (LMBM-Clust) are executed once for each $(\mathrm{dataset},k)$ pair.
Across all 13 evaluated algorithms, this amounts to $19{,}376$ total clustering executions.

All algorithms run in their most performant form. The benchmarking and plotting scripts are written in Python. Big-means-family algorithms, BDCSM, MDEClust, and DRS-means are implemented in C++; MiniBatchKMeans uses its Scikit-learn Python implementation; and LMBM-Clust, Big-Clust, and ClustSplitter run via their native Fortran libraries. The C++ and Fortran codes communicate with the main Python script through dedicated runner executables using a binary-data and JSON-result protocol.

Experiments were run in Python 3.12 on a virtual machine with 24 CPU cores and 251~GiB RAM,
hosted on an AMD EPYC 7663 server. All algorithms were executed in CPU-only
mode.

Runs that exceed the global one-hour wall-clock
limit, crash, exhaust memory, or return non-finite objectives or centers are
recorded as failures. If one run of a fixed
$(\mathrm{algorithm},\mathrm{dataset},k)$ triple exceeds the time limit, the
remaining repeats of that same triple are skipped and counted as timeout
failures, because runtimes are highly consistent for a fixed triple.

Whenever possible, competitor parameters follow the corresponding papers or
official implementations. BDCSM uses the batch-size rule
$q=\lceil \nu(\alpha) k^2/r^2\rceil$ with $\nu(0.05)=1.27359$ and
$r=0.08$, as recommended in~\citep{Alguliyev2020BDCSM}. MiniBatchKMeans uses K-means++ initialization and a batch size of 512 times the number of available cores, as recommended by the official Scikit-learn documentation to enable multi-core execution.
The nonsmooth methods use their paper-default or implementation-derived
settings. MDEClust uses a population of 150 candidate solutions and a
time-budgeted implementation ($600$ seconds per run); DRS-means uses up to $10{,}000$ swaps under the
same time-budget policy. The 600-second budget was chosen as a generous practical horizon: it gives anytime 
metaheuristics enough time for meaningful search, while keeping the full benchmark campaign over 22 datasets, 
eight values of $k$, and repeated runs computationally feasible. 
Big-means-family algorithms use K-means++ seeding or
repair with three candidate trials, local K-means tolerance $10^{-4}$, and at
most 300 local iterations per sampled problem. The original Big-means variants
use the manually optimized dataset-specific sample sizes and time budgets from~\citep{Mussabayev2023BigMeans}, with the time budget scaled linearly with $k$.

For Big-means++, the block size is fixed at $p=10$. The default
quality-oriented variant uses the convergence tolerance $q\in\{0.001,0.05\}$
and ladder depth $L=9$ according to the dataset policy; Big-means++$_\mathrm{Fast}$ uses the
more aggressive tolerance $q_{\mathrm{fast}}\in\{0.05,0.1\}$ and, on the
largest dataset, a shorter ladder $L_{\mathrm{fast}} = 5$.
For reproducibility, Table P1 of the Supplementary Material records the
parameter choices and selection rules used for every benchmark algorithm,
including paper-default competitor settings, implementation-derived settings,
and the Big-means++/Big-means++Fast tolerance and ladder policies.

\subsection{Evaluation Metrics and Aggregation}

The primary quality metric is the MSSC objective value~\eqref{eq:mssc} computed on the full dataset using the final returned centroids. To make
results comparable across datasets and values of $k$, we report relative error
with respect to the best-known objective value $f^\star$ for each
$(\mathrm{dataset},k)$ instance:
\[
E = 100\cdot\frac{f(C,X)-f^\star}{f^\star}.
\]
The best-known values are taken from the benchmark registry, which combines
values from previous MSSC studies and the best values observed in our study.

Runtime is reported in seconds. For repeated stochastic runs, the main analysis
uses the median relative error and median runtime unless explicitly stated
otherwise. The Supplementary Material expands these aggregate quantities:
Tables S1--S22 list $E_{\min}$, median $E$, $E_{\max}$, $E_{\mathrm{IQR}}$,
median runtime, $f_{\mathrm{best}}$, and failure information where applicable
for every dataset, $k$, and algorithm. Failed runs are excluded from error
statistics because they do not produce valid centroids, but their recorded or
capped runtime contributes to runtime summaries whenever timing information is
available. Figures S1--S22 provide companion per-dataset plots of central
relative error and runtime versus $k$.

In addition to final relative-error and runtime tables, the analysis scripts
produce time-to-target performance profiles, relative-error attainment profiles,
per-dataset companion plots, and representative run-variability figures.
These artifacts are used in
Section~\ref{sec:results} to separate final solution quality, computational
efficiency, anytime behavior, robustness, and the effect of multi-resolution
sample-size shaking.

The relative-error attainment profile is threshold-based. For each algorithm
and benchmark instance, the final relative error is first aggregated across
repeated runs by the median $\tilde{E}$. The profile then reports, for each
threshold $\varepsilon$, the fraction of benchmark instances on which
$\tilde{E} \le \varepsilon$. Thus, it can be read as a continuous family of
Succ@$\varepsilon$\% values.

For the time-to-target performance profile, an algorithm is credited on a
benchmark instance only when it reaches the prescribed relative-error target
$\varepsilon$. For each run, the time-to-target is the earliest wall-clock time
at which the best-so-far relative error satisfies $E \le \varepsilon$; for
methods without intermediate checkpoints, the final runtime is used if the
final solution reaches the target, and the time is treated as infinite
otherwise. Repeated runs are aggregated by the median time-to-target. Each
algorithm's median time is then divided by the best median time achieved by any
algorithm on the same instance, yielding the performance ratio shown in the
profile. Instances on which no algorithm reaches the target are excluded from
that profile.

\section{Results and Discussion}
\label{sec:results}

This section analyzes the empirical behavior of Big-means++ from three
complementary viewpoints: aggregate final solution quality and runtime,
statistical significance over the full benchmark suite, and performance-profile
behavior across relative-error and time-to-target thresholds. The central
question is whether the proposed data-native search mechanism can match or
exceed the solution quality of much heavier global-optimization hybrids while
retaining the scalability expected from a big-data clustering method.

\subsection{Aggregate Performance}

\begin{table}[t]
\centering
\caption{Global performance summary across all benchmark instances (dataset $\times$ $k$ pairs). $\bar{E}$~(\%): average relative error, computed as the mean over datasets of the mean over $k$ values of the per-$(D,k)$ median-aggregated relative error $E$; Succ@$x$\%: percentage of benchmark instances on which the algorithm's median-aggregated final relative error satisfies $E \le x\%$, shown for $x \in \{0.1, 1, 5\}$; Fail\%: average per-instance run failure rate, i.e.\ the mean across benchmark instances of the per-instance fraction of runs classified as failures (crash, OOM, timeout $>3600$\,s, or non-finite objective); $\bar{t}$~(s): average execution time, aggregated identically to $\bar{E}$; $\bar{r}_E$, $\bar{r}_t$: mean rank by error and by time across all benchmark instances (rank~1 = best; lower is better). Bold marks the best value in each column.}
\label{tab:summary}
\medskip
\begin{tabular}{lrrrrrrrrr}
\toprule
Algorithm & $\bar{E}$~(\%) & Succ@0.1\% & Succ@1\% & Succ@5\% & Fail\% & $\bar{t}$~(s) & $\bar{r}_E$ & $\bar{r}_t$ \\
\midrule
\textsc{BDCSM} & 46804.74 & 5.1 & 17.0 & 38.6 & \textbf{0.0} & 8.87 & 11.46 & \textbf{1.68} \\
\textsc{MiniBatch-KM} & 235269.99 & 5.7 & 21.6 & 54.5 & \textbf{0.0} & \textbf{1.26} & 11.39 & 2.20 \\
\textsc{Forgy-KM} & 19601.37 & 29.5 & 46.6 & 63.1 & 1.0 & 24.31 & 9.00 & 3.35 \\
\textsc{Clust-Splitter} & 57242.31 & 43.2 & 59.7 & 73.9 & 3.4 & 230.43 & 5.74 & 8.48 \\
\textsc{Big-Clust} & 129526.46 & 13.1 & 39.8 & 67.0 & \textbf{0.0} & 85.59 & 9.97 & 4.76 \\
\textsc{LMBM-Clust} & 1.33 & 52.8 & 68.8 & 86.4 & 6.2 & 244.09 & 5.28 & 9.33 \\
\textsc{MDEClust} & 14.73 & 70.5 & 79.0 & 83.5 & \textbf{0.0} & 437.51 & \textbf{3.27} & 12.08 \\
\textsc{DRS-means} & 598.99 & 64.8 & 68.8 & 76.1 & \textbf{0.0} & 455.70 & 5.50 & 12.18 \\
\textsc{Big-means-Inn} & 0.84 & 38.6 & 72.2 & 92.0 & 0.3 & 30.07 & 7.63 & 6.14 \\
\textsc{Big-means-Com} & 0.32 & 48.3 & 81.8 & 96.0 & 0.1 & 33.75 & 6.73 & 7.17 \\
\textsc{BigOptimaS3} & 132931.98 & 47.2 & 71.0 & 80.1 & 0.1 & 42.85 & 6.61 & 8.42 \\
\textsc{Big-means++} & \textbf{-0.71} & \textbf{80.1} & \textbf{94.9} & \textbf{100.0} & \textbf{0.0} & 49.31 & 3.34 & 8.11 \\
\textsc{Big-means++Fast} & -0.63 & 66.5 & 93.8 & \textbf{100.0} & \textbf{0.0} & 33.08 & 4.38 & 6.74 \\
\bottomrule
\end{tabular}
\end{table}

Table~\ref{tab:summary} shows that Big-means++ gives the strongest overall
final-quality profile. It attains the best average relative error
($\bar{E}=-0.71\%$), the highest success rate at the strictest target
Succ@0.1\% ($80.1\%$), the highest Succ@1\% ($94.9\%$), and reaches the
5\% tolerance on every benchmark instance. The negative value of $\bar{E}$ does
not imply a negative objective gap; rather, it reflects that the benchmark
registry of best-known values was improved on enough instances by the present
experiments that the aggregate error, computed relative to the registry values
available before aggregation, becomes slightly negative. In practical terms,
this is evidence that Big-means++ not only approaches the previous best-known
MSSC values, but also updates them on part of the suite.

The comparison also separates raw speed from useful speed. MiniBatch-KM and
BDCSM are the fastest methods in absolute wall-clock time, but their mean
quality ranks are near the bottom and their strict-target success rates are very
low. Conversely, the strongest optimization-oriented competitors---MDEClust,
DRS-means, LMBM-Clust, and Clust-Splitter---produce much better solutions but
require substantially more time. Big-means++ occupies the most favorable
region of this trade-off: its average runtime of $49.31$ seconds is on the
same order as the fastest scalable baselines and far below the heavy
optimization hybrids, while its solution quality is better than the state of
the art in the aggregate. The efficiency-oriented Big-means++$_\mathrm{Fast}$
variant further illustrates the tunability of the method: by using more
aggressive universal settings for the probe tolerance $q$ and, on the largest
dataset, a shorter sample-size ladder, it reduces the average runtime to
$33.08$ seconds while preserving Succ@5\%$=100.0\%$ and nearly the same
average error. Thus, the universal parameters of Big-means++ do not merely
remove dataset-specific tuning; they provide a simple and interpretable
quality--time control for practitioners.

Figures S24 and S25 in the Supplementary Material summarize error and CPU-time
categories separately for small-$k$ and large-$k$ regimes, complementing
Table~\ref{tab:summary} with a count-based view of how often each algorithm
falls into practically meaningful performance bands.

\subsection{Statistical Comparison}

\begin{table}[!ht]
\centering
\caption{Statistical comparison of \textsc{Big-means++} versus baseline algorithms over 176~benchmark instances (dataset, $k$ pairs). Per-instance errors are median-aggregated across runs ($\tilde{E}_{A,i}$). \textbf{Rank}: average rank in the Friedman omnibus test (rank~1 = lowest error; computed on complete-case instances where all algorithms have results). \textbf{$\bm{N}$}: instances where both the target and the competitor have finite aggregated errors. \textbf{W/T/L}: wins~/ ties~/ losses for the target (win if $\tilde{E}_{\mathrm{target}} < \tilde{E}_{\mathrm{comp}} - \delta$, with $\delta=10^{-2}$ percentage points of relative error; loss if reversed; otherwise tie). \textbf{$\bm{W}$}: Wilcoxon signed-rank test statistic $\min(W^+,W^-)$ (two-sided; $N$ common instances). \textbf{$\bm{p}_{\mathrm{Holm}}$}: Holm--Bonferroni corrected $p$-value ($^*p{<}0.05$, $^{**}p{<}0.01$, $^{***}p{<}0.001$). \textbf{Med~$\bm{\Delta}$}: median of ${\tilde{E}}_{\mathrm{comp},i} - {\tilde{E}}_{\mathrm{target},i}$ over $N$ common instances (positive~=~target is better, in \%). \textbf{95\%~CI}: bootstrap percentile interval for Med~$\Delta$ (10\,000~resamples).}
\label{tab:statistical}
\vspace{4pt}
\noindent\textbf{Friedman test} (165~complete-case instances, 13~algorithms): $\chi^2_F = 1075.76$, $p < 0.001$.\\[2pt]
\noindent\textbf{Target algorithm:} \textsc{Big-means++}, avg.~rank~$= 3.38$.

\medskip

\begin{tabular}{l r r r r r l}
\toprule
Competitor & Rank & $N$ & W\,/\,T\,/\,L & $W$ & $p_{\mathrm{Holm}}$ & $\mathrm{med}\,\Delta$ (95\%\,CI) \\
\midrule
\textsc{BDCSM} & 11.58 & 176 & 173\,/\,3\,/\,0 & 2.0 & $<\!0.001$$^{***}$ & +12.45 ([+6.57, +22.42]) \\
\textsc{MiniBatch-KM} & 11.52 & 176 & 170\,/\,6\,/\,0 & 5.0 & $<\!0.001$$^{***}$ & +3.85 ([+2.90, +5.82]) \\
\textsc{Forgy-KM} & 9.13 & 176 & 145\,/\,30\,/\,1 & 165.0 & $<\!0.001$$^{***}$ & +1.32 ([+0.66, +2.64]) \\
\textsc{Clust-Splitter} & 5.74 & 170 & 95\,/\,63\,/\,12 & 2561.0 & $<\!0.001$$^{***}$ & +0.28 ([+0.00, +0.59]) \\
\textsc{Big-Clust} & 10.01 & 176 & 176\,/\,0\,/\,0 & 0.0 & $<\!0.001$$^{***}$ & +1.88 ([+1.29, +2.74]) \\
\textsc{LMBM-Clust} & 5.28 & 165 & 78\,/\,73\,/\,14 & 3391.0 & $<\!0.001$$^{***}$ & +0.00 ([+0.00, +0.22]) \\
\textsc{MDEClust} & 3.09 & 176 & 51\,/\,93\,/\,32 & 7590.0 & $0.973$ & -0.00 ([-0.00, -0.00]) \\
\textsc{DRS-means} & 5.10 & 176 & 58\,/\,83\,/\,35 & 6217.0 & $0.054$ & -0.00 ([-0.00, +0.00]) \\
\textsc{Big-means-Inn} & 7.80 & 176 & 153\,/\,23\,/\,0 & 25.0 & $<\!0.001$$^{***}$ & +0.25 ([+0.12, +0.55]) \\
\textsc{Big-means-Com} & 7.02 & 176 & 136\,/\,23\,/\,17 & 1304.0 & $<\!0.001$$^{***}$ & +0.10 ([+0.05, +0.14]) \\
\textsc{BigOptimaS3} & 6.91 & 176 & 144\,/\,20\,/\,12 & 977.0 & $<\!0.001$$^{***}$ & +0.09 ([+0.05, +0.13]) \\
\bottomrule
\end{tabular}%
\end{table}

Table~\ref{tab:summary} aggregates all algorithms, whereas the statistical
comparison in Table~\ref{tab:statistical} asks a stricter paired question: on the same
$(\mathrm{dataset},k)$ instances, does Big-means++ systematically improve on
each competitor? The Friedman omnibus test rejects equality of ranks across the
13 algorithms ($\chi^2_F=1075.76$, $p<0.001$), confirming that the observed
performance differences are not random rank fluctuations. Big-means++ has an
average Friedman rank of $3.38$, placing it among the leading methods while
also being much faster than the other high-quality competitors.

The pairwise results show clear and statistically significant advantages over
the classical scalable baselines, the nonsmooth methods Clust-Splitter,
Big-Clust, and LMBM-Clust, and all previous Big-means-family methods. The
largest margins occur against methods whose speed comes at the expense of
solution quality: for example, Big-means++ wins on 173 of 176 common instances
against BDCSM and on 170 against MiniBatch-KM. More importantly for the
algorithmic contribution, Big-means++ also improves strongly over its direct
predecessors: 153/23/0 against Big-means-Inn, 136/23/17 against
Big-means-Com, and 144/20/12 against BigOptimaS3. These gains support the
combined effect of the flowing incumbent, multi-resolution sample-size shaking, and
the automatic probe-based convergence mechanism. The probe-based detector is
important because it regulates each flowing trajectory autonomously: difficult
trajectories continue while meaningful landscape-driven movement remains,
whereas exhausted trajectories stop before additional perturbations consume
resources or degrade their terminal states.

For MDEClust and DRS-means the interpretation is more nuanced. Both are strong
global-search competitors, and many paired differences fall within the
practical-tie band. The tie threshold is set to $0.01$ percentage points of
relative error, which is safely above floating-point accumulation noise after
normalization but still far below a difference that would be actionable in
practice. Under this rule, Big-means++ records 51 wins, 93 ties, and 32 losses
against MDEClust, and 58 wins, 83 ties, and 35 losses against DRS-means. Thus,
Big-means++ is best described as practically competitive with these strongest
hybrid or swap-based rivals, while still winning slightly more often than it
loses. The Wilcoxon statistic $W=\min(W^+,W^-)$ should be read together with
the corrected $p$-value: it measures the smaller of the positive and negative
signed-rank sums after pairing instances, so very small values indicate that
almost all nonzero signed ranks favor the same method. For $N=176$, the total
rank mass is $1+2+\cdots+176=15576$; under the null hypothesis of no systematic
difference, one would expect roughly half of this mass on each side, i.e.,
$W^+\approx W^-\approx 7788$. This is why $W=0$ for
Big-Clust and $W=25$ for Big-means-Inn correspond to highly significant
advantages. In contrast, the much larger $W$ values for MDEClust and
DRS-means, together with Holm-corrected $p$-values of $0.973$ and $0.054$,
indicate that their paired signed-rank evidence is not sufficient to claim a
statistically significant median difference after correction, despite the
favorable win/loss counts for Big-means++.

\subsection{Performance Profiles and Anytime Behavior}

\par\bigskip
\noindent\begin{minipage}{\linewidth}
    \centering
    \begin{minipage}[t]{0.48\linewidth}
        \centering
        \includegraphics[width=\linewidth]{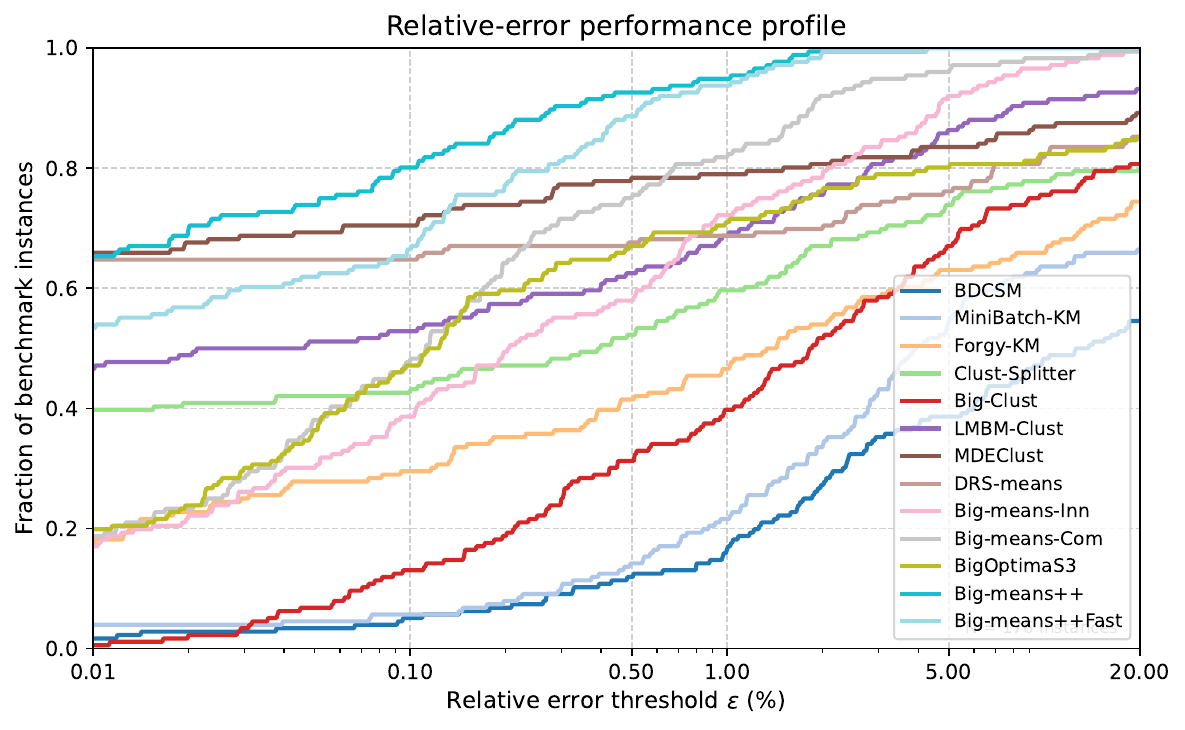}
    \end{minipage}\hfill
    \begin{minipage}[t]{0.48\linewidth}
        \centering
        \includegraphics[width=\linewidth]{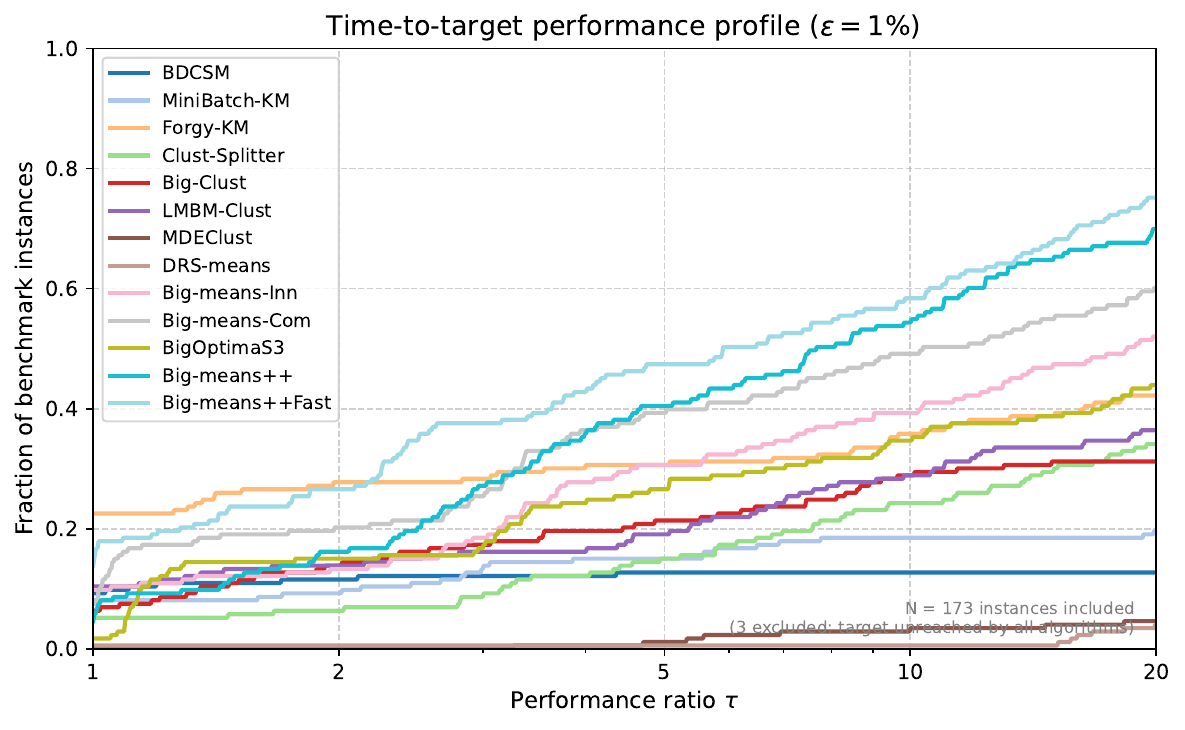}
    \end{minipage}
    \captionof{figure}{Performance profiles across all 176 benchmark instances (dataset~$\times$~$k$ pairs). \textbf{(a) Left:} Relative-error attainment profile --- the $x$-axis is the relative-error threshold $\varepsilon$, and each curve gives the fraction of instances on which the algorithm's median final relative error $\tilde E$ over repeated runs satisfies $\tilde E \le \varepsilon$. \textbf{(b) Right:} Time-to-target profile at $\varepsilon = 1\%$ --- the $x$-axis is the performance ratio $\tau$, i.e.\ the algorithm's median time-to-target divided by the best median time-to-target on the same instance; each curve gives the fraction of instances reached within that ratio. Higher and further-left curves indicate better performance.}
    \label{fig:profiles}
\end{minipage}
\par\bigskip

Figure~\ref{fig:profiles} provides a threshold-wise view of the same
phenomenon. In the relative-error attainment profile, Big-means++ is the
dominant curve over the most important low-error region. At the strict
$0.1\%$ threshold it solves roughly four fifths of the benchmark instances,
whereas even the strongest alternatives remain visibly below it. The
Big-means++$_\mathrm{Fast}$ curve follows closely, confirming that a moderate
relaxation of the universal quality-time parameters preserves most of the
benefit of the full algorithm. The gap is especially pronounced against
classical scalable methods: their curves rise only at much looser error
thresholds, showing that their speed is obtained by accepting substantially
shallower local minima.

The time-to-target profile at $\varepsilon=1\%$ reinforces the same
conclusion from a computational perspective. Big-means++ reaches the strict
target on the largest final fraction of included instances, while Big-means++$_\mathrm{Fast}$ reaches a large subset earlier. Methods such as
MDEClust and DRS-means can be very accurate on some instances, but their curves
remain low in this profile because they need much more time to reach the same
strict target. In contrast, the fastest baselines appear early but plateau at
low fractions, reflecting limited attainable quality. This combination is the
main practical advantage of Big-means++ for big-data MSSC: it behaves like a
scalable sampling-based method in runtime, but like a high-quality global
optimization method in final objective value.
Figure S23 in the Supplementary Material gives companion time-to-target
profiles for stricter $0.1\%$ and looser $5\%$ thresholds, showing how this
speed-quality trade-off changes across accuracy levels.

To complement the aggregate profiles, Figure~\ref{fig:anytime} reports
representative anytime trajectories on four benchmark instances with $k=25$.
The instances were chosen to cover distinct empirical regimes: Skin
Segmentation represents low-dimensional clustering with many points; ISOLET
represents high-dimensional clustering; EEG Eye State represents a smaller and
cluster-imbalanced case; and US Census Data 1990 represents the very
large-scale setting with moderate dimensionality.

\par\bigskip
\noindent\begin{minipage}{\linewidth}
    \centering
    \includegraphics[width=\linewidth]{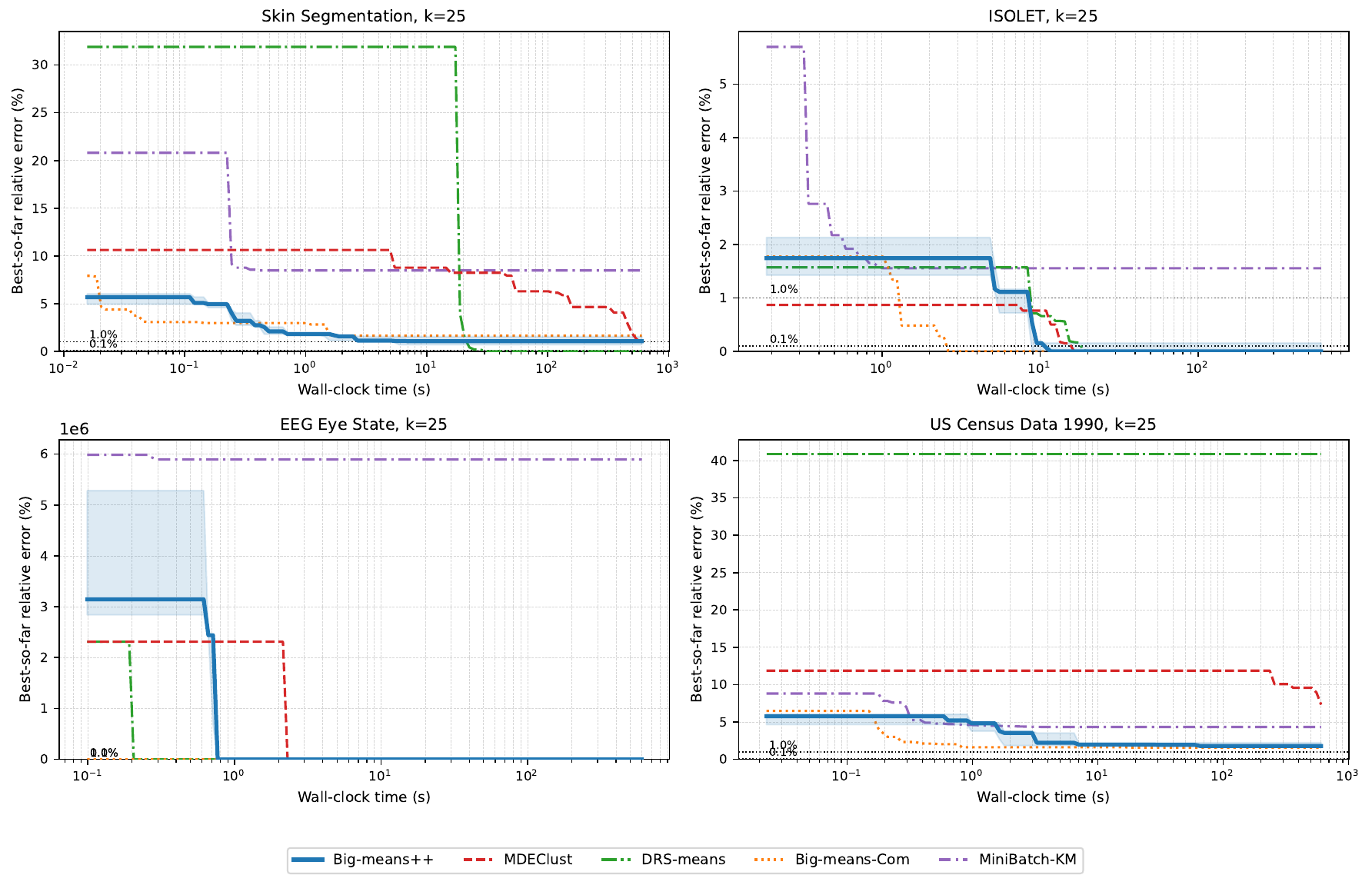}
    \captionof{figure}{Representative anytime behavior on four benchmark instances with $k=25$. Each curve reports the median best-so-far relative error across five executions of the corresponding algorithm. The shaded band around the \textsc{Big-means++} curve shows its interquartile range across the same five executions.}
    \label{fig:anytime}
\end{minipage}
\par\bigskip

The anytime curves show that Big-means++ improves rapidly across all four
representative regimes and continues to push the best-so-far error toward the
strict target levels as time increases. The shaded interquartile band indicates
that this behavior is not driven by a single lucky run, but is stable across
independent executions. This is especially important for big-data clustering,
where an algorithm must not only reach strong final objectives, but also make
reliable progress under finite wall-clock budgets.
Additional run-to-run variability evidence is given in Figure S26 of the
Supplementary Material, which uses violin/box plots for representative datasets
at $k=5$ and $k=25$ and annotates failed-run counts where they occur.

\subsection{Sensitivity Analysis}

The sensitivity analysis reveals a clear trade-off between solution quality and execution time as $p$ increases. Table~\ref{tab:bigmeanspp-p-sensitivity} shows that the smallest setting, $p=1$, is fast but substantially degrades robustness and solution quality, producing a very large average relative error and succeeding within $5\%$ on only $68.8\%$ of the instances. Increasing $p$ from $5$ to $50$ progressively improves the average error and error rank, although at a corresponding runtime cost. The largest setting, $p=200$, achieves the lowest average error ($0.31\%$), but requires $301.21$ seconds on average---more than fourteen times the runtime of the default setting, $p=10$---while providing only modest improvements in the success rates.

Overall, $p=10$ offers a favorable balance between accuracy and computational cost. This supports its use as the default setting in the main experiments.

\begin{table}[t]
\centering
\caption{Sensitivity of \textsc{Big-means++} to the parameter $p$ on
the four datasets used in the anytime analysis. Results cover all
32 benchmark instances (four datasets $\times$ eight values of $k$).
For US Census Data 1990, each instance was repeated five times; all
other instances were repeated ten times. The reported quantities are
aggregated as in Table~\ref{tab:summary}. The setting $p=10$ is the
default used in the main experiments. Bold marks the best value in
each column among the tested settings.}
\label{tab:bigmeanspp-p-sensitivity}
\medskip
\begin{tabular}{crrrrrrrrr}
\toprule
$p$ & $\bar{E}$~(\%) & Succ@0.1\% & Succ@1\% & Succ@5\%
    & Fail\% & $\bar{t}$~(s) & $\bar{r}_E$ & $\bar{r}_t$ \\
\midrule
1   & 114039.66 & 21.9 & 40.6 & 68.8  & \textbf{0.0} & \textbf{5.25} & 8.03 & \textbf{5.28} \\
5   & 0.46      & 50.0 & \textbf{84.4} & \textbf{100.0}
    & \textbf{0.0} & 13.77 & 4.22 & 6.84 \\
10 (default) & 0.40 & 59.4 & 81.2 & \textbf{100.0}
    & \textbf{0.0} & 20.89 & 3.81 & 7.72 \\
50  & 0.34 & \textbf{62.5} & \textbf{84.4} & \textbf{100.0}
    & \textbf{0.0} & 79.54 & \textbf{3.47} & 9.38 \\
200 & \textbf{0.31} & \textbf{62.5} & \textbf{84.4}
    & \textbf{100.0} & \textbf{0.0} & 301.21 & 3.72 & 11.00 \\
\bottomrule
\end{tabular}
\end{table}

\subsection{Ablation Study}

\begin{table}[t]
\centering
\caption{Ablation summary for the main Big-means++ ingredients across all 176 benchmark instances. Each ablated variant disables one structural component while keeping the remaining protocol unchanged. For \textsc{Big-means++-Ladder} and \textsc{Big-means++-Conv}, the fixed sample size and maximum runtime, respectively, are manually optimized per dataset and coincide with the optimal values used in~\citep{Mussabayev2023BigMeans}. Metrics are aggregated as in Table~\ref{tab:summary}. Bold marks the best value in each column.}
\label{tab:ablation}
\medskip
\setlength{\tabcolsep}{3pt}
\begin{tabular}{llrrrrrr}
\toprule
Variant & Ablated component & $\bar{E}$~(\%) & Succ@0.1\% & Succ@1\% & Succ@5\% & Fail\% & $\bar{t}$~(s) \\
\midrule
\textsc{Big-means++} & none & \textbf{-0.71} & \textbf{80.1} & \textbf{94.9} & \textbf{100.0} & \textbf{0.0} & 49.31 \\
\textsc{Big-means++-IncFlow} & flowing incumbent & 1568.98 & 74.4 & 93.8 & 99.4 & \textbf{0.0} & 70.77 \\
\textsc{Big-means++-Ladder} & sample-size ladder & -0.59 & 68.2 & 90.9 & 97.2 & 4.8 & 137.49 \\
\textsc{Big-means++-Conv} & convergence detector & 1977.55 & 73.3 & 89.2 & 96.0 & \textbf{0.0} & \textbf{40.33} \\
\bottomrule
\end{tabular}
\end{table}

Table~\ref{tab:ablation} isolates the contribution of the three main
Big-means++ ingredients. The full algorithm is best on every quality metric:
lowest average error, highest strict-target success, highest 1\% and 5\%
attainment, and zero failures. Removing the flowing-incumbent mechanism is the
least damaging ablation by success rate, but it still reduces strict-target
attainment and increases runtime, indicating that unconditional centroid
movement improves both search mobility and efficiency. Removing the
sample-size ladder causes the clearest scalability and reliability loss, even
though the resulting fixed sample size is manually optimized for each dataset
using values from~\citep{Mussabayev2023BigMeans}: the variant becomes much slower,
introduces failures, and loses substantial strict-target success. Because the
ablation already uses optimized fixed sizes, these losses cannot be attributed
merely to the absence of automatic hyperparameter selection. Instead, they
support the algorithmic role of dynamic sample-size variation: traversing
multiple resolution scales broadens the family of surrogate landscapes explored
and improves the resulting search trajectories. Removing the
convergence detector gives the fastest ablation, but at a clear quality cost,
despite using the manually optimized per-dataset maximum runtimes from the same
study. Thus, its benefit cannot be explained merely by automatic budget
selection or computational savings. The detector is an optimization driver: it
regulates where each flowing trajectory terminates, preserving strong terminal
states before unnecessary continuation can turn useful mobility into harmful
drift. At the same time, it allocates computation adaptively and removes manual
per-dataset budget tuning. Thus, the
advantage of Big-means++ is structural rather than accidental: its performance
comes from the interaction of flowing search, multi-resolution landscape shaking,
and convergence regulation.

Taken together, the tables and profiles support the data-native optimization
claim of the paper. Big-means++ does not rely on recombination, random swaps,
bundle-method machinery, or repeated full-data passes during search. Instead,
it obtains global-search behavior by organizing the input-curation modality:
the flowing incumbent transfers information between sample-induced landscapes,
the geometric ladder shakes the search across multiple resolutions and broadens
the family of landscapes traversed, autonomous agents preserve complementary
trajectories, common-landscape competition transforms their diversity into a
collective decision, and the probe-based convergence
detector terminates each agent near the end of its productive search phase,
protecting terminal quality while adapting computational effort to trajectory
difficulty. The empirical result is a
method that is robust, parameter-light, and especially suitable for clustering
large datasets where both full-data global search and low-quality fast
approximations are unsatisfactory.

\FloatBarrier

\section{Conclusion, Limitations and Future Work}
\label{sec:conclusion}

Big-means++ shows that high-quality global search for big-data MSSC does not
require a heavy metaheuristic layer or repeated full-data optimization. The key
idea is to turn data sampling itself into the search mechanism: flowing centroids
move through sample-induced landscapes, the sample-size ladder shakes this
trajectory across multiple resolutions and broadens the landscape family
explored, a competitive population of autonomous agents supplies collective
trajectory diversity, and the convergence detector stops each agent before
productive movement gives way to unnecessary or potentially harmful
continuation, while simultaneously providing adaptive resource allocation and
universal speed-quality control. Across 176 benchmark instances, this simple data-native design
achieves state-of-the-art final quality with practical runtimes, zero failures,
strong anytime behavior, and ablation evidence showing that its advantage comes
from the interaction of its main components rather than from a single isolated
trick.

The main limitation is that Big-means++ remains a heuristic: it does not provide
a certificate of global optimality, and the best-known objective values used for
evaluation are not necessarily true optima. The study is also restricted to
Euclidean MSSC with prescribed values of $k$ and numerical datasets.

Future work will extend this data-native approach beyond the input-curation
optimization modality alone. In particular, we plan to combine adaptive data
sampling with systematic changes in the objective formulation and more explicit
landscape-exploration mechanisms through hybridization with metaheuristics.
Such multimodal designs may further improve global optimization capability and
robustness while preserving the scalability and simplicity that make
Big-means++ effective for big-data clustering.

\section*{Supplementary Material}

Supplementary material associated with this article will be available online at
the Pattern Recognition article page. It contains detailed benchmark parameter
policies, complete per-dataset numerical tables, companion per-dataset figures,
additional time-to-target profiles, small-$k$ and large-$k$ category summaries,
and run-to-run variability analyses.

\section*{CRediT authorship contribution statement}

\textbf{Ravil Mussabayev:} Conceptualization, Methodology, Software,
Validation, Formal analysis, Investigation, Resources, Data curation,
Writing -- original draft, Writing -- review \& editing, Visualization,
Supervision, Project administration, Funding acquisition.
\textbf{Rustam Mussabayev:} Conceptualization, Validation, Resources,
Writing -- review \& editing.
\textbf{Zukhra Yerdaliyeva:} Software, Validation, Visualization.
\textbf{Kuldeyev Nursultan:} Software, Validation, Visualization.

\section*{Acknowledgements}

This research was funded by the Science Committee of the Ministry of Science and Higher Education of the Republic of Kazakhstan (grant no. AP26197157).

\section*{Declaration of Interests}

The authors declare that they have no known competing financial interests or
personal relationships that could have appeared to influence the work reported
in this paper.

\section*{Declaration of generative AI and AI-assisted technologies in the manuscript preparation process}

During the preparation of this work, the authors used ChatGPT for language
editing. The authors reviewed and edited the output as needed and take full
responsibility for the content of the published article.

\bibliographystyle{elsarticle-num}

\bibliography{references}

\end{document}